\documentclass[12pt]{spieman}  
\usepackage{amsmath,amsfonts,amssymb}
\usepackage{graphicx}
\usepackage{setspace}
\usepackage{tocloft}

\title{FSDNet-An efficient fire detection network for complex scenarios based on YOLOv3 and DenseNet}

\author[a]{Li Zhu}
\author[a]{Jiahui Xiong}
\author[b]{Wenxian Wu}
\author[a,b,*]{Hongyu Yu}
\affil[a]{Nachang University, Cvip Group}

\cftpagenumbersoff{figure}
\cftpagenumbersoff{table} 
\begin{document} 
\maketitle

\begin{abstract}
Fire is one of the common disasters in daily life. To achieve fast and accurate detection of fires, this paper proposes a detection network called FSDNet (Fire Smoke Detection Network), which consists of a feature extraction module, a fire classification module, and a fire detection module. Firstly, a dense connection structure is introduced in the basic feature extraction module to enhance the feature extraction ability of the backbone network and alleviate the gradient disappearance problem. Secondly, a spatial pyramid pooling structure is introduced in the fire detection module, and the Mosaic data augmentation method and CIoU loss function are used in the training process to comprehensively improve the flame feature extraction ability. Finally, in view of the shortcomings of public fire datasets, a fire dataset called MS-FS (Multi-scene Fire And Smoke) containing 11938 fire images was created through data collection, screening, and target annotation. To prove the effectiveness of the proposed method, the accuracy of the method was evaluated on two benchmark fire datasets and MS-FS. The experimental results show that the accuracy of FSDNet on the two benchmark datasets is 99.82$\%$ and 91.15$\%$, respectively, and the average precision on MS-FS is 86.80$\%$, which is better than the mainstream fire detection methods.
\end{abstract}



\begin{spacing}{2}   

\section{Introduction}
\label{sect:intro}  
Fires are characterized by suddenness and spread, and can cause huge economic losses and seriously threaten human life safety in a short period of time. According to data released by the Fire and Rescue Bureau of the National Emergency Management Department in 2022, there were a total of 748,000 reported fires in China in 2021, causing 1,987 deaths, 2,225 injuries, and direct property losses of up to 6.75 billion yuan. Therefore, researching fire automatic detection methods that combine real-time and accuracy has important theoretical and practical significance.

Generally speaking, fire detection methods can be roughly divided into two types: sensor-based detection methods\cite{ko2009fire} and vision-based detection methods\cite{geetha2021machine} . Sensor-based detection methods played a crucial role in early fire detection, which mainly sense the characteristic changes accompanying a fire, such as heat, light, and smoke concentration, through sensors\cite{gaur2019fire} . However, the fire features from the source require a certain amount of time to be sensed by deployed sensors, which leads to the response of corresponding sensors being lagging. This characteristic also limits the applicability of this method to large spaces or public places. In addition, this method also requires human and other equipment involvement to further determine information such as the size, location, and degree of burning of the fire\cite{kim2019video} .

With the popularization of digital cameras and the rapid development of various technologies such as image processing, machine learning, and deep learning, video-based fire detection methods have been rapidly developed. This type of method captures real-time video image information indoors and outdoors through cameras, transmits the images to a data processing center, performs real-time processing and analysis, and finally judges the fire situation on-site and takes corresponding response measures. In the early days, vision-based methods were accomplished by manually extracting single or multiple features from fire images, including static features such as brightness, color, and texture, and dynamic features such as flicker and motion\cite{pundir2019dual} . For example, Celik et al.\cite{celik2009fire} effectively detected fires from video sequences by combining the YCbCr color space model with foreground object information. Chen et al.\cite{chen2004early} proposed a method that combines the RGB color space model with dynamic analysis to detect flames and uses the growth characteristics of the flames to perform secondary judgment on suspected areas. However, detection methods based on a single static feature are difficult to adapt to various complex situations and are easily affected by environmental factors.

To overcome this limitation, researchers have considered combining dynamic features with the static characteristics of flames. Dedeoglu et al.\cite{dedeoglu2005real} proposed a flame detection method that combines color, motion features, and wavelet domain analysis. Time-domain wavelet transforms are used to determine the boundaries of flames, while spatial-domain wavelet transforms are used to detect changes in flame color. Toreyin et al.\cite{toreyin2005flame} utilized an implicit Markov model to detect the flickering process of flames, effectively reducing false alarms caused by dynamic flame color similar objects, but the method is sensitive to changes in lighting conditions. Habiboglu et al.\cite{habiboglu2011real} developed a video-based wildland fire detection system that obtains the moving area of flames through background subtraction and color thresholding, and then extracts relevant features from the motion area. Zhao et al.\cite{zhao2011svm} constructed static SVM classifiers and dynamic SVM classifiers based on 11 static features and 27 dynamic features of fire samples to achieve real-time detection of forest fires, but the false detection rate was high.

Most of the methods proposed by the aforementioned researchers are based on hand-crafted features and have limitations in their application scenarios. Therefore, they have gradually been replaced by methods based on deep learning\cite{hinton2006reducing} . Deep learning-based methods can more effectively extract deep features and semantic information from fire images, further improving the performance of fire detectors.

In recent years, there have been many fire detection methods based on Convolutional Neural Networks (CNNs)\cite{albawi2017understanding} . Frizzi et al.\cite{frizzi2016convolutional} used a CNN for fire and smoke detection and tested it with real fire video sequences, showing superior classification performance compared to traditional fire detection methods. Gonzalez et al.\cite{gonzalez2017accurate} proposed a fire detection model combining CNNs and unmanned aerial vehicles based on AlexNet\cite{krizhevsky2017imagenet} and convolutional sequences, but the model had a high false alarm rate. In\cite{hongchao2021behavior}, a video fire detection method was proposed by combining Faster R-CNN with LSTM to extract spatial and temporal features of flames. In\cite{gaur2020video}, a video fire detection method that simultaneously considers dynamic features based on motion flicker and depth static features was proposed. Some researchers have applied CNNs, Transformers\cite{ghali2021wildfire}, and image segmentation\cite{minaee2021image} to the field of visual fire detection to develop trained models for collecting fire image features. Sharma et al.\cite{sharma2017deep} used VGG16 and ResNet50 to perform fire detection in images and provided a self-collected fire dataset. Zhang et al.\cite{wu2018using} compared the detection performance of forest fires using object detection algorithms such as Faster R-CNN\cite{ren2015faster}, YOLO\cite{redmon2016you,redmon2017yolo9000,redmon2018yolov3} , and SSD\cite{liu2016ssd}, where SSD showed better real-time performance and detection accuracy. They further optimized the network structure and proposed a new Tiny-YOLO network, demonstrating its effectiveness.

Although some achievements have been made in the aforementioned fire detection methods, the overall accuracy is still low, and they have limited applicability, making it difficult to achieve high robustness in complex real-life scenarios. To address these issues, this paper proposes a fire detection network named FSDNet, which combines classification and detection functions based on YOLOv3 and is used for fire monitoring and prevention. Our main contributions can be summarized as follows:

(1) A fire detection network named FSDNet based on YOLOv3 and dense connections is proposed. The network consists of a feature extraction module, a fire classification module, and a fire detection module, which combines classification and detection functions and can perform real-time and accurate fire recognition and detection on image and video data. 

(2) In the FSDNet network, a dense connection structure is introduced in the feature extraction module, which reduces the number of parameters while strengthening feature propagation through module stacking, small-scale convolution kernels, and convolutional pooling. The fire detection branch incorporates spatial pyramid pooling to fuse global and local features. The CIOU loss function is used as the bounding box regression loss to improve the precision of predicted bounding box localization.

(3) To address the insufficiency of publicly available fire datasets, we independently established a fire dataset called MS-FS, which contains 11,938 fire images. Through a large amount of material collection, data screening, and object annotation work, the dataset covers hundreds of real fire scenarios.

The experimental results show that the proposed FSDNet network has higher accuracy than existing mainstream fire detection methods. The overall accuracy, false alarm rate, and miss rate on two benchmark fire datasets DS1 and DS2 are 99.82$\%$, 0.07$\%$, 0.17$\%$ and 91.15$\%$, 10.28$\%$, 7.56$\%$, respectively. The mean Average Precision (mAP) on MS-FS reaches 86.80$\%$, which is 8.5$\%$, 6.4$\%$, and 1.2$\%$ higher than YOLOv3, Gaussian-YOLOv3, and YOLOv4\cite{bochkovskiy2020yolov4}, respectively. It can accurately and real-time accomplish fire detection tasks. The proposed FSDNet will have a promising application prospect in community fire monitoring and prevention.

\section{Method}

\begin{figure}
\begin{center}
\begin{tabular}{c}
\includegraphics[height=12cm]{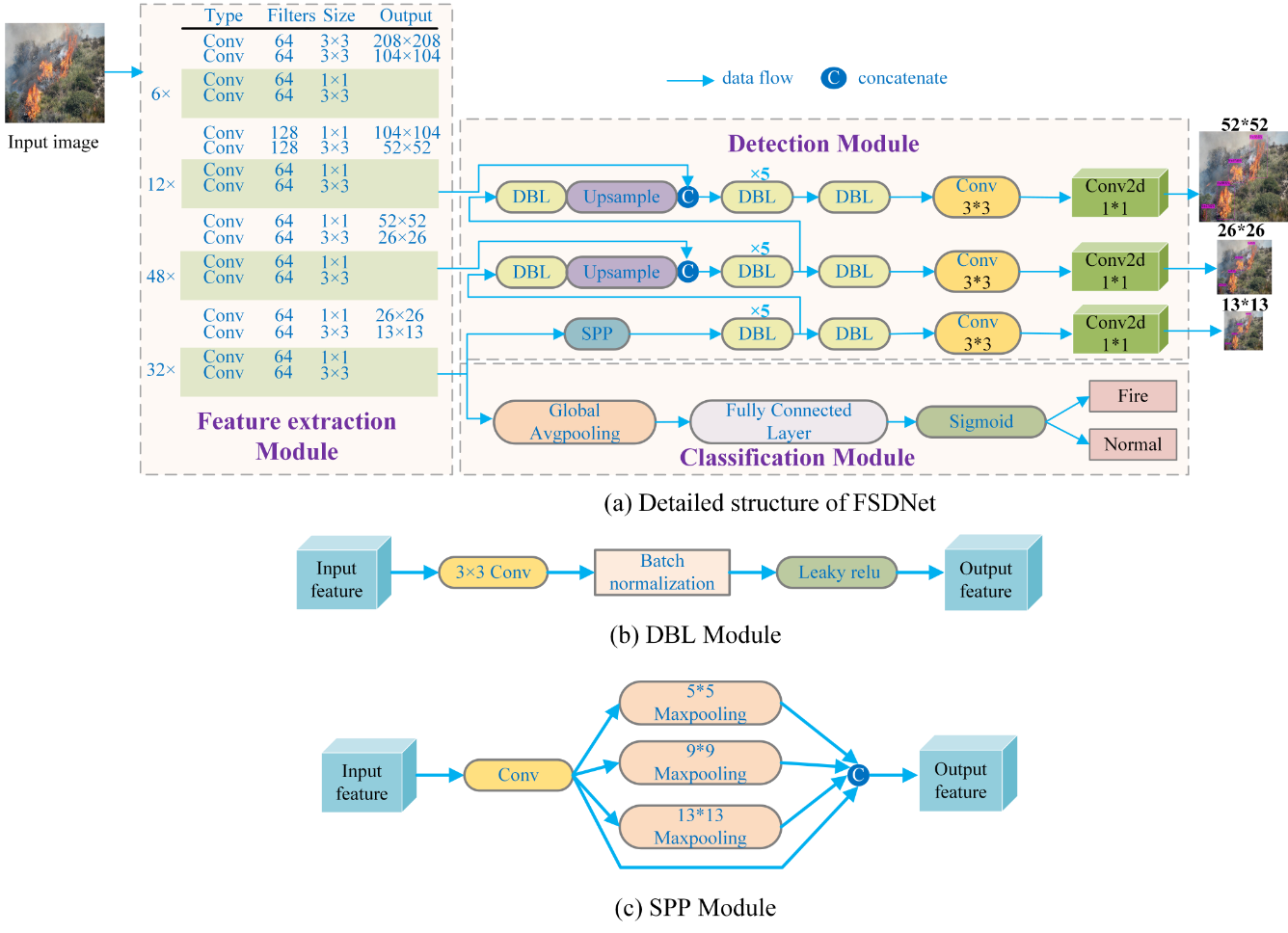}
\end{tabular}
\end{center}
\caption 
{ \label{fig:ex1}
FSDNet algorithm framework diagram. } 
\end{figure}

As shown in Fig.~\ref{fig:ex1}, the overall framework of the proposed fire detection network FSDNet consists of three modules: a dense-connected basic feature extraction module, a fire classification module, and a fire detection module with three detection branches. This network takes the flame dataset created in this paper as input and establishes two models with the same feature extraction module, respectively used to complete the fire classification task and real-time fire detection task.

\subsection{Feature Extraction Module}
\label{sect:title}

\begin{figure}
\begin{center}
\begin{tabular}{c}
\includegraphics{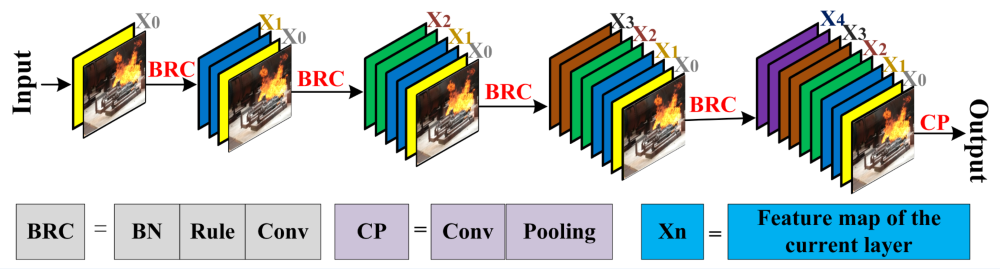}
\end{tabular}
\end{center}
\caption 
{ \label{fig:ex2}
Dense connection structure in the basic feature extraction module. } 
\end{figure}

DenseNet\cite{huang2017densely} has shown excellent performance in classification experiments based on CIFAR and ImageNet datasets, achieving lower error rates with fewer parameters compared to ResNet\cite{he2016deep}. The specific connectivity structure of the dense connections is shown in Fig.~\ref{fig:ex2}. By establishing dense connections such that the input of each layer is derived from the output of all preceding layers, feature reuse is achieved in the channel dimension, maximizing information exchange between preceding and succeeding layers. This connectivity method has the following advantages: 1. Dense connections are equivalent to connecting each layer directly to the input and loss, effectively mitigating the problem of gradient vanishing. 2. Connecting all preceding layers to succeeding layers enhances feature propagation. 3. Compared to the common residual connection method of ResNet, the dense connection method significantly reduces the number of parameters.

Based on the sensitivity of error rates to fire detection and the advantages of DenseNet, FSDNet chose the dense connection method used in DenseNet for the convolutional layer connections in the basic feature module. The introduced dense connection feature extraction module consists of four convolutional layers, three pooling layers, three Dense Block modules, and one fully connected layer, which increases the feature reuse, making it highly meaningful for detecting objects like fire and smoke that do not have fixed shapes. Additionally, the feature extraction module uses small convolution kernels for feature extraction, consisting only of small convolution kernels (1*1, 3*3), to obtain a deeper and wider network with fewer parameters. 

The features extracted during the initialization stage are fed into the first Dense Block, while adjacent Dense Blocks are connected by 1*1 convolutional layers and max pooling layers for transition. The 1*1 convolutional layer is used for dimension reduction, and the max pooling layer suppresses local image noise and reduces interference from background items. The 1x1 convolutional layer acts as the bottleneck layer, reducing the computational load by decreasing feature dimensions and merging features from different channels. The 3*3 convolutional layer captures information from the eight-neighborhood of pixels, with the smallest kernel size, and cascading multiple 3*3 convolutional layers can capture the same receptive field with fewer parameters. For example, replacing one 7*7 convolutional layer with three cascading 3*3 convolutional layers can reduce 45$\%$ of computational load and parameters. Additionally, the stacking of nonlinear activation functions increases the network's ability to perform nonlinear fitting.

\subsection{Fire Classification Module}

As shown in Fig.~\ref{fig:ex1}(a), after passing through the basic feature extraction module, the different layer feature maps generated by the input image are separately transmitted to the classification module and detection module.

To fully utilize the global information of the input, the classification module uses a global average pooling layer\cite{lin2013network} to receive the feature maps generated by the final output of the basic feature extraction module. The global average pooling layer can collect the spatial information of the input feature maps to obtain the global context information of the object; at the same time, its introduction can complete regularization throughout the entire network structure, reduce the number of parameters, and effectively prevent overfitting.After the global average pooling layer, a fully connected layer is connected to it, which can complete the learning of global information for the corresponding feature map, and can selectively emphasize useful features and suppress useless features. Considering that fire detection is a binary classification task, the activation function of the output layer is selected as Sigmoid. The final output of this module is Fire/Normal.

\subsection{Fire Detection Module}

The YOLOv3 detection framework was applied to the fire detection module, which shows great advantages in accuracy and real-time, suitable for fire detection tasks. To alleviate the problem of vanishing gradient that is easily occurred in the feature extraction network Darknet53 in YOLOv3, the feature extraction network proposed in section 2.1 of this paper is selected as the Backbone. Its dense connection structure makes full use of the feature information of different layers in the process of feature forward propagation and significantly reduces the number of network parameters. In addition, this paper further improves YOLOv3 in terms of model training and structure optimization, aiming to complete the fire detection task accurately and in real-time.

\subsubsection{Architechture}

Similar to the idea of FPN\cite{lin2017feature}, the detection part adopts a multi-scale detection method, where the feature maps output by the 24th, 44th, and 204th layers of the feature extraction module are respectively passed to the three detection branches of the detection part. The small feature map, with a larger receptive field and richer semantic information, generates small-scale detectors responsible for detecting large-sized objects. In addition, the small feature map, after upsampling, is concatenated with the feature map from the 44th layer of the feature extraction module to obtain medium-sized detectors responsible for detecting medium-sized object s. Similarly, the medium-sized feature map, after upsampling, is concatenated with the feature map from the 24th layer of the feature extraction module to obtain large-sized feature maps with higher resolution and accurate position information, and the large-scale detectors are responsible for detecting small-sized objects.

\subsubsection{Spatial Pyramid Pooling Module and Mosaic}

\begin{figure}
\begin{center}
\begin{tabular}{c}
\includegraphics{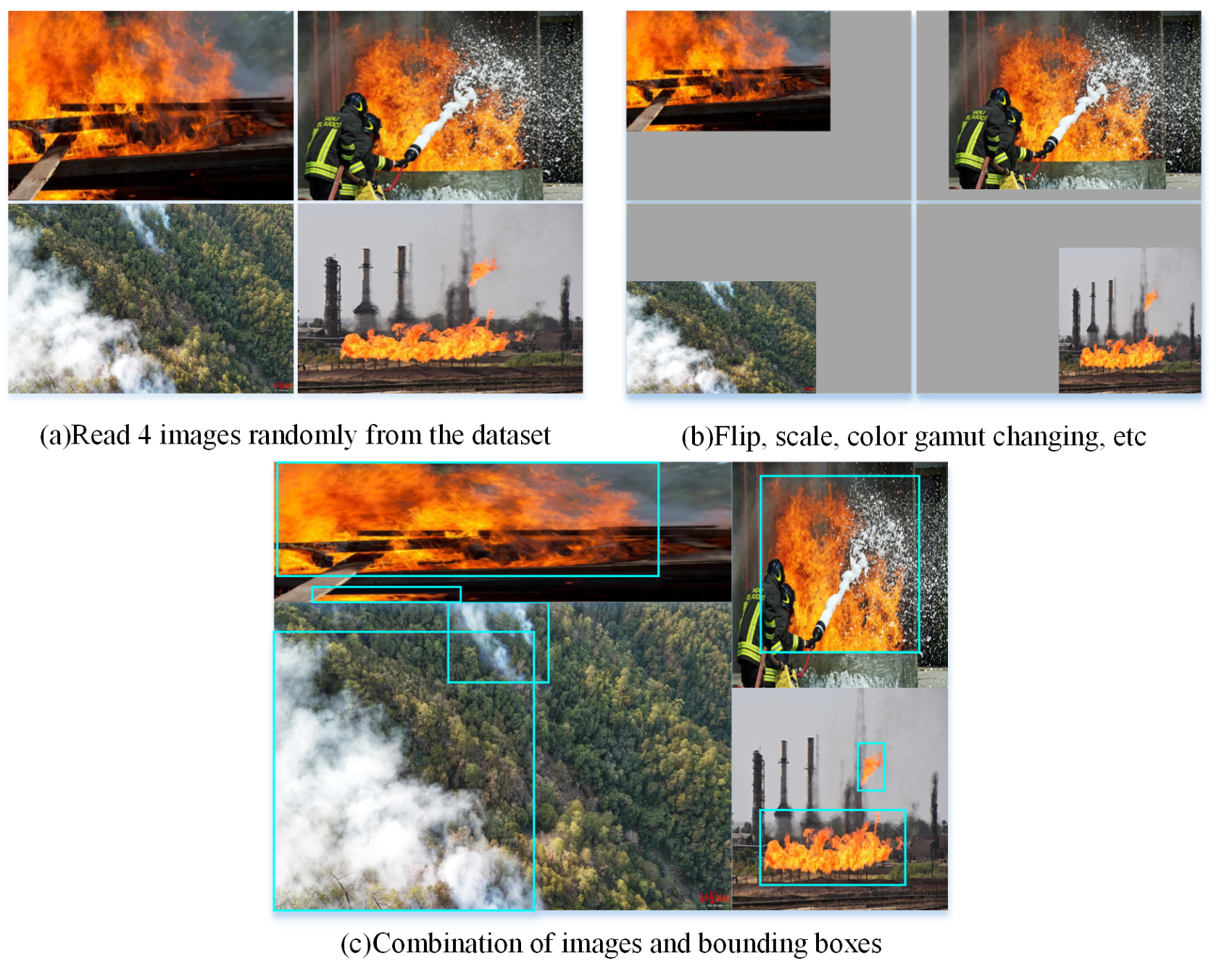}
\end{tabular}
\end{center}
\caption 
{ \label{fig:ex3}
The realization process of the Mosaic data enhancement method. } 
\end{figure}

Due to the morphological characteristics and large differences in size of fire detection objects (flames, smoke), a spatial pyramid pooling structure\cite{he2015spatial} is added to reduce the accuracy loss caused by object morphological differences in detection. The main purpose of the spatial pyramid pooling structure, first proposed, is to solve the following two problems: 1. image distortion caused by operations such as clipping and scaling of the input image; 2. repetitive extraction of image features by convolutional neural networks.

In this paper, the SPP module is introduced between the basic feature extraction module and the bottom detection branch of the detection module. The SPP module consists of four parallel branches, with kernel sizes of 5*5, 9*9, 13*13 max pooling layers and a skip connection. The largest pooling kernel size in the SPP module is designed to be as consistent as possible with the size of the input feature map, in order to strengthen the fusion between features of different sizes. The outputs of the four branches are concatenated at the feature map level, which enables the fusion of global and local features in the feature extraction module, greatly enriches the expression ability of the feature map, improves the recognition accuracy, and is beneficial for detecting objects with large differences in size, such as flames and smoke.

Considering the unbalanced proportion of objects with different sizes in the MS-FS established in section 3 of this paper, we use Mosaic during training. Mosaic is a data augmentation technique first proposed in YOLOv4, which randomly scales, crops, and arranges four images and feeds the resulting image into the training network, equivalent to training with four images at once. This approach not only speeds up the training process but also greatly enriches the background of detected objects.

The specific implementation process of Mosaic is as follows: firstly, four images are randomly selected from the dataset; secondly, each original image is subjected to operations such as flipping, scaling, color space transformation (brightness, saturation, hue), and the images are placed in order from top-left to top-right; finally, adjustments and combinations of images and object candidate boxes are made. As shown in Fig.~\ref{fig:ex3}.

\subsubsection{Loss Function}
The complete form of the loss function for the detection algorithm FSDNet proposed in this article is shown as follows:
\begin{equation}
\label{eq:fov}
Loss = {L_{box}} + {L_{obj}} + {L_{cls}}\\
\end{equation}

\begin{equation}
\label{eq:fov}
         = {\lambda _{coord}}\sum\limits_{i = 0}^{{K^{\rm{2}}}} {\sum\limits_{j = 0}^M {I_{ij}^{obj}(2 - {w_i} \times {h_i})} } [{({x_i} - \mathop {{x_i}}\limits^ \wedge  )^2} + {({y_i} - \mathop {{y_i}}\limits^ \wedge  )^2}]\\
\end{equation}

\begin{equation}
\label{eq:fov}
           + {\lambda _{coord}}\sum\limits_{i = 0}^{{K^{\rm{2}}}} {\sum\limits_{j = 0}^M {I_{ij}^{obj}(2 - {w_i} \times {h_i})} } [{({w_i} - \mathop {{w_i}}\limits^ \wedge  )^2} + {({h_i} - \mathop {{h_i}}\limits^ \wedge  )^2}]\\
\end{equation}    

\begin{equation}
\label{eq:fov}
           - \sum\limits_{i = 0}^{{K^{\rm{2}}}} {\sum\limits_{j = 0}^M {I_{ij}^{obj}} } [\mathop {{C_i}}\limits^ \wedge  \log ({C_i}) + (1 - \mathop {{C_i}}\limits^ \wedge  )\log (1 - {C_i})]\\
\end{equation}

\begin{equation}
\label{eq:fov}
           - {\lambda _{noobj}}\sum\limits_{i = 0}^{{K^{\rm{2}}}} {\sum\limits_{j = 0}^M {I_{ij}^{noobj}} } [\mathop {{C_i}}\limits^ \wedge  \log ({C_i}) + (1 - \mathop {{C_i}}\limits^ \wedge  )\log (1 - {C_i})]\\
\end{equation}

\begin{equation}
\label{eq:fov}
           - \sum\limits_{i = 0}^{{K^{\rm{2}}}} {I_{ij}^{obj}} \sum\limits_{c \in class} {[\mathop {{p_i}}\limits^ \wedge  (c)\log ({p_i}(c)) + (1 - \mathop {{p_i}}\limits^ \wedge  (c))\log (1 - {p_i}(c))]} \\
\end{equation}

In this paragraph, $K^2$ represents the total number of grids into which the input image is divided, and $M$ represents the number of candidate boxes generated by each grid. Each candidate box will produce a corresponding bounding box after passing through the network, and the total number of bounding boxes obtained will be $K^2\times M$ . $\lambda_{coord}$ and $\lambda_{noobj}$ are hyperparameters that represent the weights of the coordinate error and class error, respectively. $I_{ij}^{noobj}$ represents whether the $j$ th anchor box of the $i$ th grid is responsible for the object object. If it is responsible, then $I_{ij}^{noobj}$ is 1; otherwise, it is 0. The responsibility is specifically defined as follows: among all anchor boxes of the $j$ th anchor box in the $i$ th grid, if the IoU (Intersection over Union) with the ground truth of the object is the highest, then the anchor box is responsible for predicting the object. Similarly, $I_{ij}^{noobj}$ represents that the $j$ th anchor box of the $i$ th grid is not responsible for the object object.

In YOLOv3, the Smooth L1 Loss function is responsible for the position loss, which overcomes the drawbacks of both MAE and MSE. However, treating (x, y, w, h) as independent variables separates their relative relationship and ignores their holistic nature. Therefore, this article introduces CIoU as the algorithm's bounding box regression loss. CIoU is an improvement on DIoU loss, which not only considers the overlap between two bounding boxes and the distance between their centers but also adds an adjustment parameter. Based on this, the aspect ratio of the bounding box is also considered, which is more favorable for detecting objects with large differences in shape and size, such as flames and smoke.

The formula for the CIoU loss function introduced in this article is shown below:
\begin{equation}
\label{eq:fov}
{L_{box}} = {L_{CIoU}} = 1 - IoU + \frac{{{\rho ^2}(b,{b^{gt}})}}{{{c^2}}} + \alpha v
\end{equation}

Among them, $b$ represents the center point of the predicted bounding box, $\rho$ represents the center point of the ground truth bounding box, $\rho$ denotes the Euclidean distance between the center points of the two bounding boxes, $c$ is the diagonal distance of the smallest rectangle that can simultaneously cover the two bounding boxes. $v$ is responsible for measuring the similarity of aspect ratios, $\alpha$ is the weight coefficient, and the specific functional forms of both are shown in the formula.

\begin{equation}
\label{eq:fov}
v = \frac{4}{{{\pi ^2}}}{(\arctan \frac{{{\omega ^{gt}}}}{{{h^{gt}}}} - \arctan \frac{\omega }{h})^2}
\end{equation}
\begin{equation}
\label{eq:fov}
\alpha  = \frac{v}{{(1 - IoU) + v}}
\end{equation}

In the above equation, $h^{gt}$ and $\omega ^{gt}$ represent the height and width of the ground truth bounding box, while $h$ and $\omega$ represent the height and width of the predicted bounding box, and $IoU$ denotes the intersection over union between the two boxes.

\section{Establishment of Fire and Smoke Dataset}
Deep learning models have a certain degree of dependency on datasets, and a high-quality and properly partitioned dataset can accelerate model convergence speed and improve model performance. Therefore, this article first investigated and summarized the existing publicly available fire and smoke datasets. Table~\ref{tab:table1} shows a comparison between the dataset created in this article and some of the existing publicly available fire datasets.


\begin{table}[ht]
\caption{Comparison with other fire dataset.}
\label{tab:table1}
\begin{center}
\begin{tabular}{c c c c }
\hline
\begin{tabular}[c]{@{}c@{}}Dataset\\ source\end{tabular} & Content Description                                                                                                                                       & \begin{tabular}[c]{@{}c@{}}Object\\ Classes\end{tabular} & \begin{tabular}[c]{@{}c@{}}Pixel level\\ Annotation\end{tabular} \\ \hline
Bowfire dataset                                          & \begin{tabular}[c]{@{}c@{}}An image dataset including\\ 119 fire images\\ and 106 normal images.\end{tabular}                                             & Fire                                                     & False                                                            \\ \hline
Bilkent University                                       & \begin{tabular}[c]{@{}c@{}}A video dataset composed of 14\\ fire videos and 17 normal videos.\\ The average duration of the videos\\ is142s.\end{tabular} & Fire                                                     & False                                                            \\ \hline
The University of Modena                                 & \begin{tabular}[c]{@{}c@{}}A video dataset consisting of 14\\ outdoor smoke videos.The\\ average duration of the videos is\\ 12.6s.\end{tabular}          & Smoke                                                    & False                                                            \\ \hline
Keimyung University                                      & \begin{tabular}[c]{@{}c@{}}A video dataset including 16\\ indoor and outdoor fire videos.\\ The average duration of the videos\\ is 39s.\end{tabular}     & Fire\&Smoke                         & False                                                            \\ \hline
The proposed dataset                                     & \begin{tabular}[c]{@{}c@{}}A dataset composed of 11938 fire\\ images with preprocessing,\\ covering hundreds of real-time\\ fire scenarios.\end{tabular}  & Fire\&Smoke                         & True                                                             \\ \hline
\end{tabular}
\end{center}
\end{table}

The research results show that existing publicly available datasets have shortcomings such as small data volume, limited coverage of scenarios, imbalanced data, and lack of annotation information. Therefore, this article has created a large-scale flame and smoke dataset, MS-FS, with annotation information. The detailed process is as follows.

(1) Data collection. A web crawler was built to search and download fire-related videos and images. Ultimately, 70 video clips of real fire scenes and thousands of fire images were collected, covering complex fire scenarios in daily life as much as possible, such as city roads, large buildings, supermarkets, and parking lot fires.

(2) Video frame splitting. The collected video clips were processed by frame splitting, dividing each second of video into 25 frames. To avoid excessive similarity between consecutive images, at least one image was extracted as valid data every 3 frames.

(3) Data filtering. In order to ensure the validity of the data, the data was screened according to the following criteria. First, according to the principle of object proportion, filter out the data whose area of flame and smoke in each scene accounts for less than 30$\%$; secondly, according to the principle of image signal-to-noise ratio, filter out the data whose signal-to-noise ratio is less than or equal to 35 dB; finally, according to theRGB color model, filter out the data whose background RGB value conforms to the following formula:
\begin{equation}
\label{eq:fov}
R  >  {R_{avg}}\\
\end{equation}
\begin{equation}
\label{eq:fov}
G  >  {G_{avg}}\\
\end{equation}
\begin{equation}
\label{eq:fov}
R  >  G  >  {\rm B}
\end{equation}

(4) Object annotation. Correct category labeling of images is extremely important for the dataset. The selected data was annotated one by one using the image annotation tool, LabelImg. The annotated objects were divided into two categories: flames and smoke. The data format for annotation was the XML file format of the standard dataset, Pascal VOC. This file contained annotation information for the original image, including the object position, object category, and bounding box position information. 

MS-FS contains 11,938 images covering hundreds of real fire scenes, including various backgrounds from simple to complex, environments from indoors to outdoors, lighting from day to night, and heights from ground level to high-rise buildings. MS-FS effectively solves the problems of small data volume, limited coverage of scenarios, imbalanced data, and lack of annotation information in existing publicly available datasets. MS-FS is divided into three categories, which are scenes containing only flames, scenes containing only smoke, and scenes containing both flames and smoke. The annotation examples for these three types of data are shown in Figure 4, where the purple and yellow boxes represent the annotations for smoke and flames, respectively.

\begin{figure}
\begin{center}
\begin{tabular}{c}
\includegraphics{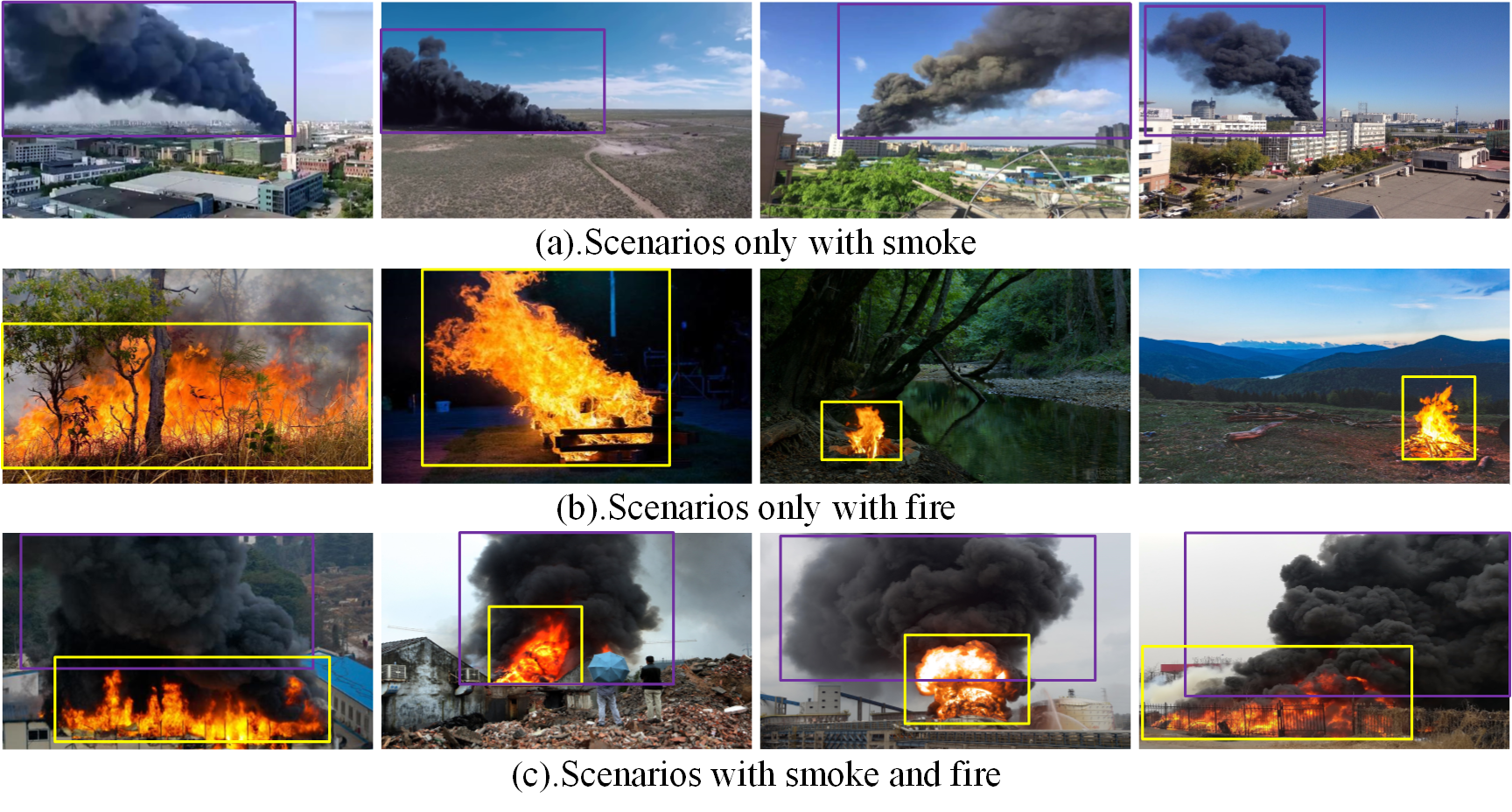}
\end{tabular}
\end{center}
\caption 
{ \label{fig:ex4}
Labeled examples of the dataset proposed in this paper. } 
\end{figure}

\section{Experiment and Result}

\subsection{Dataset Description}
In order to validate that the proposed FSDNet algorithm can still achieve excellent performance in different scenarios and to compare it more intuitively with other fire detection algorithms, this study selected DS1\cite{foggia2015real} and DS2\cite{chino2015bowfire} as the test set and some of the training set for the FSDNet classifier, and chose the dataset MS-FS proposed in this paper as the training and test set for the FSDNet detector.

DS1 is a fire video dataset consisting of 31 video clips captured by cameras in different scenarios, with a total duration of 74 minutes (4450 seconds). Among the 31 video clips, the first 14 contain frames with fires, while the remaining 17 do not and were captured in normal environments. This dataset imposes a high demand on the model's performance, as it contains scenes with heavy smoke, fire-colored objects, and fire objects at different distances.

DS2 consists of 119 fire images and 107 non-fire images, making it a highly challenging dataset that is commonly used to test the detection accuracy of fire detection methods.
 
MS-FS is a self-prepared fire dataset in this paper.

\subsection{Experimental Results}

\subsubsection{Classification results of FSDNet}
This paper uses three commonly used evaluation metrics to assess the performance of the FSDNet classifier and compare it with other methods: false positive rate (FP Rate), false negative rate (FN Rate), and accuracy. A lower FP Rate and FN Rate and a higher accuracy indicate better classifier performance. In fire detection tasks, FP Rate represents the fire false alarm rate, while FN Rate represents the fire miss rate.

In the experiments based on DS1, this study compared several relevant methods, including those based on handcrafted feature extraction and deep learning, and the comparison results are shown in Table~\ref{tab:table2}.

\begin{table}[ht]
\caption{Comparison with other fire detection methods on DS1.} 
\label{tab:table2}
\begin{center}       
\begin{tabular}{c c c c} 
\hline\hline
\rule[-1ex]{0pt}{3.5ex}  Method & FP Rate & FN Rate & Accuracy \\
\hline\hline
\rule[-1ex]{0pt}{3.5ex}  Rafiee et al. (RGB)\cite{rafiee2011fire} & 41.18 & 7.14 & 74.20 \\
\rule[-1ex]{0pt}{3.5ex}  Celik et al.& 29.41 & \textbf{0} & 83.87 \\
\rule[-1ex]{0pt}{3.5ex}  Chen et al.& 11.76 & 14.29 & 87.10 \\
\rule[-1ex]{0pt}{3.5ex}  Habiboglu er al.& 5.88 & 14.29 & 90.32 \\
\rule[-1ex]{0pt}{3.5ex}  Lascio et al.\cite{lascio2014improving} & 6.67 & \textbf{0} & 92.59\\
\rule[-1ex]{0pt}{3.5ex}  Foggia et al.& 11.76 & \textbf{0} & 93.55\\
\rule[-1ex]{0pt}{3.5ex}  Muhammad et al.\cite{muhammad2018efficient} & 8.87 & 2.12 & 94.50\\
\rule[-1ex]{0pt}{3.5ex}  FSDNet & \textbf{0.07} & 0.17 & \textbf{99.82} \\
\hline\hline
\end{tabular}
\end{center}
\end{table} 

(1) The false positive rate (FPR) of fire alarms is discussed in this article. The FSDNet algorithm proposed in this study has the lowest FPR of only 0.07$\%$, which is 5.81$\%$ to 41.11$\%$ lower than other methods. This suggests that the FSDNet algorithm proposed in this study is the most reliable and robust. The second-lowest FPR is Habiboglu's method, with a rate of 5.88$\%$. However, this method has a higher false negative rate (FNR) of 14.29$\%$, which is the highest value among all the methods in Table~\ref{tab:table2}.

(2) The false negative rate (FNR) was studied and the methods proposed by Lascio, Foggia, and Celik showed the best performance, achieving zero FNR on the DS1 test set. However, these three algorithms had high false positive rates of 6.67$\%$, 11.76$\%$, and 29.41$\%$ respectively. On the other hand, FSDNet exhibited some false positives but still maintained a relatively low FNR of 0.17$\%$, making it a superior performing algorithm.

(3) The accuracy of fire detection was studied, and FSDNet achieved the highest accuracy among all methods with a score of 99.82$\%$, approaching 100$\%$, and obtaining outstanding results. Its accuracy was 5.32$\%$ higher than the second-ranked algorithm proposed by Muhammad.

In summary, among the three performance metrics compared, the FSDNet algorithm proposed in this paper showed the highest accuracy (99.82$\%$), the lowest false positive rate (0.07$\%$), and excellent false negative rate (0.17$\%$). Its overall performance in fire detection was the best among all methods in Table~\ref{tab:table2}.

\begin{figure}
\begin{center}
\begin{tabular}{c}
\includegraphics{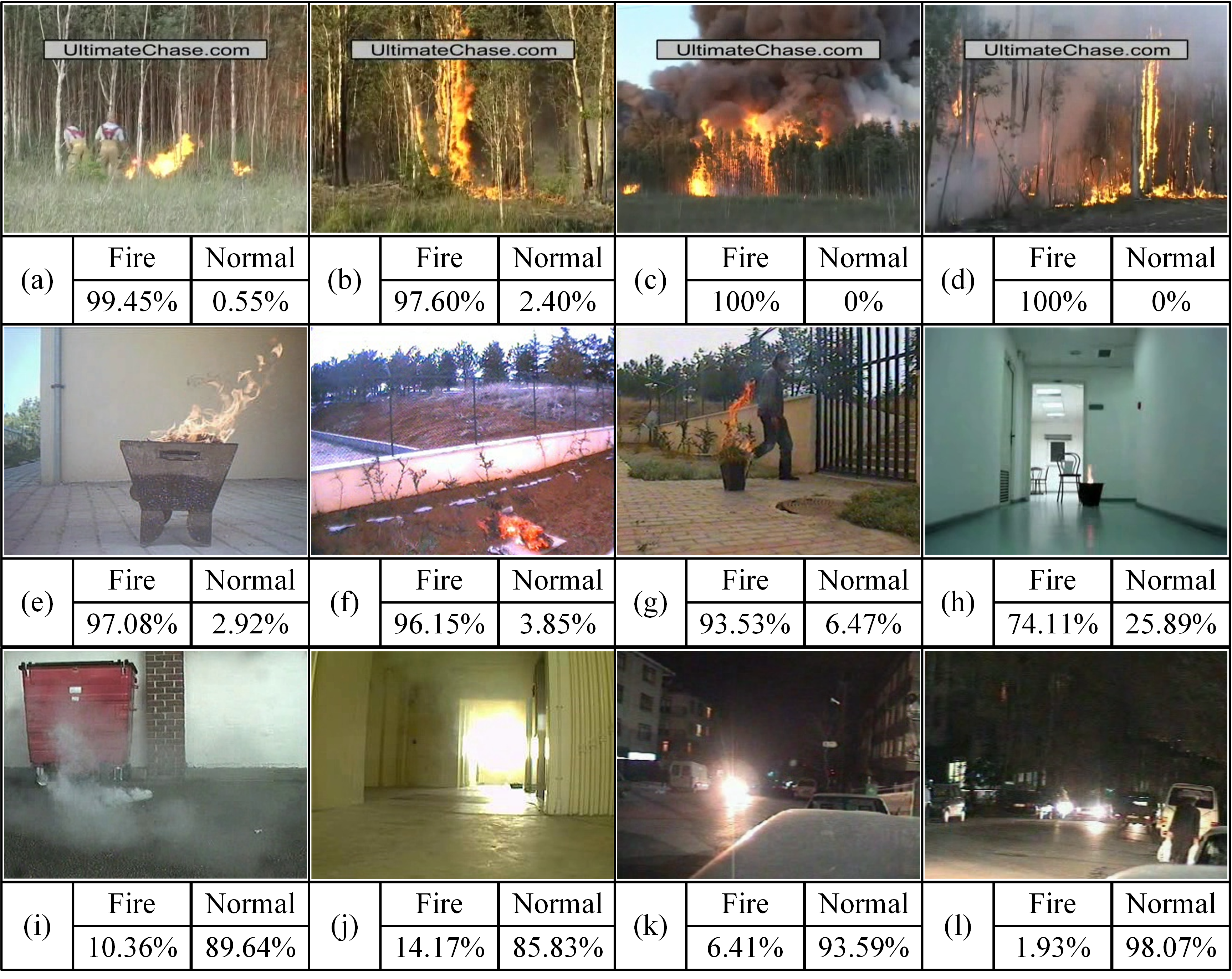}
\end{tabular}
\end{center}
\caption 
{ \label{fig:ex5}
Fire classification results On DS1 of FSDNet. (a)-(d)Conventional fire scenarios with obvious flames. (e)-(h)Single fire source scenario with a relatively small object. (i)-(l)Scenarios containing interferences items.} 
\end{figure}

The fire classification results of FSDNet based on DS1 are shown in Fig.~\ref{fig:ex5}.


From the fire results in Fig.~\ref{fig:ex5}, it can be seen that:

(1) Even in the face of conventional fire scenes with clearly visible flame objects such as (a)-(d), FSDNet can achieve a detection accuracy of up to 99$\%$ even if the shape of the flame object varies significantly. This means that FSDNet can easily handle routine fire detection tasks.

(2) When faced with a single fire source scene with a small object area such as the similar scenes (e)-(h), although the proportion of the fire source area gradually decreases from the scene (e)-(h), the The detection results of FSDNet are always kept at a reliable level ($>$90$\%$). Even if the flame object is relatively small, such as scene (h), the algorithm also guarantees an accuracy of 74$\%$, which shows that FSDNet can be 
effectively applied to the task of early fire detection. By detecting a small area of flame in the early stage of fire, the fire spread can be curbed, reduce the loss caused by fire.

(3) In the face of non-fire scenarios with certain interference items such as similar scenarios (i)-(l), the detection accuracy rate of FSDNet also remains above 90$\%$, which is also the embodiment of the low false alarm rate in Table~\ref{tab:table2}. Among them, scene (i) is moving smoke, scene (j) is an indoor light source, and scenes (k) and (l) are both outdoor light sources at night. This shows that FSDNet has good anti-jamming performance.

From the above three sets of fire detection results, it can be concluded that FSDNet maintains a high detection accuracy when faced with conventional fire scenes, fire scenes with a relatively small flame area, and non-fire scenes with some interference. The algorithm demonstrates strong resistance to interference and shows potential for application in early fire detection tasks.

The performance of the FSDNet algorithm is also compared with five existing high-performance, lightweight object detection algorithms MobileNet, ShuffleNet, ShuffleNetV2, MobileNetV2 and SqueezeNet on the DS2 dataset. The results are shown in Table~\ref{tab:table3}.

\begin{table}[ht]
\caption{Comparison with other lightweight networks on DS2. } 
\label{tab:table3}
\begin{center}       
\begin{tabular}{c c c c} 
\hline\hline
\rule[-1ex]{0pt}{3.5ex}  Method & FP Rate & FN Rate & Accuracy \\
\hline\hline
\rule[-1ex]{0pt}{3.5ex}  MobileNet & 23.36  & 12.61 & 82.30 \\
\rule[-1ex]{0pt}{3.5ex}  ShuffleNet & 21.50 & 10.92 & 84.07 \\
\rule[-1ex]{0pt}{3.5ex}  ShuffleNetV2 & 16.82 & 9.24 & 87.17 \\
\rule[-1ex]{0pt}{3.5ex}  MobileNetV2 & 15.09 & 9.32 & 87.95 \\
\rule[-1ex]{0pt}{3.5ex}  SqueezeNet & 14.15 & \textbf{6.72} & 89.78\\
\rule[-1ex]{0pt}{3.5ex}  FSDNet & \textbf{10.28} & 7.56 & \textbf{91.15} \\
\hline\hline
\end{tabular}
\end{center}
\end{table} 

From Table~\ref{tab:table3}, it can be concluded that FSDNet has the lowest false positive rate (FPR) among all methods, which is 10.28$\%$; SqueezeNet shows the best false negative rate (FNR) at 6.72$\%$, but its false positive rate is as high as 14.15$\%$, which is 3.87$\%$ higher than that of FSDNet. In terms of accuracy, FSDNet still performs the best with an accuracy of 91.15$\%$. Overall, FSDNet achieves the best combination of the three performance indicators, showing the lowest false positive rate and the highest accuracy. Considering that the DS2 dataset covers a wide range of scenes and has high challenges in terms of interference, the excellent performance of FSDNet on this test set can to some extent reflect its generalization ability in complex real-world situations.

\begin{figure}
\begin{center}
\begin{tabular}{c}
\includegraphics{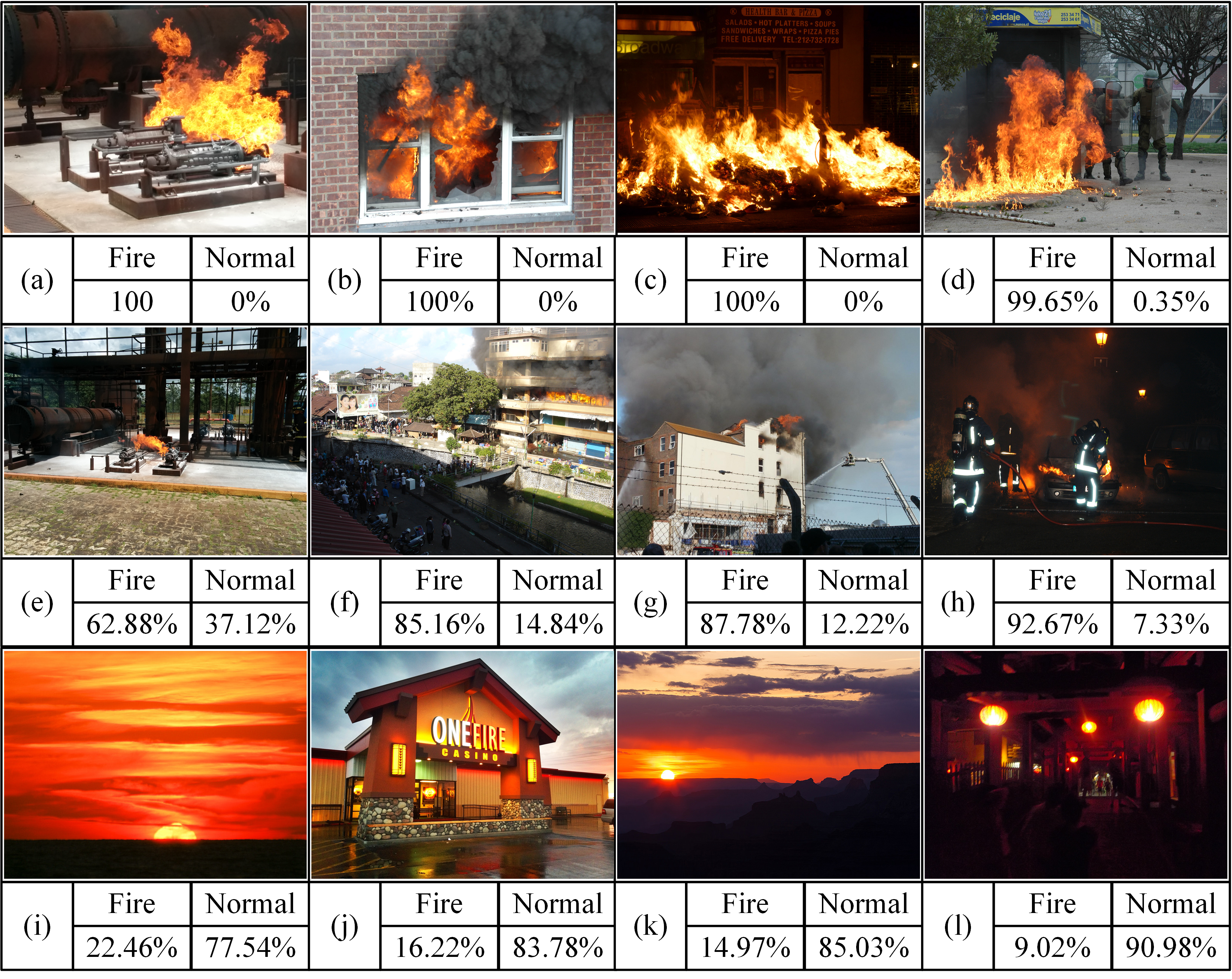}
\end{tabular}
\end{center}
\caption 
{ \label{fig:ex6}
Fire classification results On DS2 of FSDNet. (a)-(d)Common fire scenarios occurred in daily life (e)-(h)Fire scenarios that are more complex and challenging. (i)-(l)Normal scenarios with extremely disturbing objects.} 
\end{figure}

An example of the detection results of FSDNet on the DS2 test set is shown in Fig.~\ref{fig:ex6}. From the fire detection results in Figure 6, it can be seen that:

(1) Common fire scenarios. Such as scenarios (a)-(d), the algorithm in this paper shows excellent performance of 100$\%$ accuracy and 99.65$\%$ accuracy.

(2) Complex fire scene. For example, in scene (e), the object area of the flame is small; in scene (f), the fire occurs in a building and has obvious flame color and shape characteristics; scene (g), the object area of the flame is small and covered by smoke ; Scenario (h), street lights appear as interference items and small fire sources at the same 
time, and the accuracy of FSDNet remains above 80$\%$, which shows that FSDNet has the ability to handle fire detection tasks in complex environments.

(3) Non-fire scenarios with strong interference items. For example, in scenarios (i)-(l), the interference items are sunset, dim indoor light, morning glow and night light. FSDNet also guarantees an accuracy rate higher than 85$\%$ and has strong antiinterference ability.

In general, the proposed fire detection algorithm FSDNet shows that the FSDNet has high accuracy, low false alarm rate and low false alarm rate in the face of common fire scenes, complex fire scenes and non-fire scenes with strong interference items. The false alarm rate can be better able to perform the task of fire detection.

\subsubsection{Detection results of FSDNet}

To verify the effectiveness of the FSDNet detector, FSDNet was compared with other deep learning algorithms, including the original YOLOv3, YOLOv3-dense, and the high-performance YOLOv4. All experiments in this section were based on the MS-FS dataset proposed in this paper, and the Pascal VOC's mean average precision (mAP) calculation method was used as the evaluation metric to assess the performance of each method. The experimental results are shown in Table~\ref{tab:table4}.

\begin{table}[ht]
\caption{Comparison with other fire detection methods on DS3. } 
\label{tab:table4}
\begin{center}       
\begin{tabular}{c c c c c} 
\hline\hline
\rule[-1ex]{0pt}{3.5ex}  Method & Backbone & Fire & Smoke & mAP \\
\hline\hline
\rule[-1ex]{0pt}{3.5ex}  YOLOv3-dense & Densenet201  & 0.685 & 0.711 & 0.698 \\
\rule[-1ex]{0pt}{3.5ex}  YOLOv3 & Darknet53 & 0.775 & 0.791 & 0.783 \\
\rule[-1ex]{0pt}{3.5ex}  YOLOv4 & CSPResNext50 & 0.853 & \textbf{0.858} & 0.856 \\
\rule[-1ex]{0pt}{3.5ex}  FSDNet & Feature Extraction Module & \textbf{0.934} & 0.801 & \textbf{0.868}\\
\hline\hline
\end{tabular}
\end{center}
\end{table} 

It can be seen from Table~\ref{tab:table4} that the mAP of the FSDNet algorithm on the DS3 dataset is higher than that of other algorithms, which is 86.8$\%$, which is 17$\%$, 8.5$\%$ and 1.2$\%$ higher than that of YOLOv3-dense, YOLOv3 and YOLOv4, respectively. In particular, the detection accuracy of FSDNet for flame category objects is significantly improved, which is 24.9$\%$ and 15.9$\%$ higher than YOLOv3-dense and YOLOv3. Even compared to the high-performance YOLOv4, the improvement is as high as 8.1$\%$, which means that the FSDNet algorithm has better fire detection performance and can detect flame objects more accurately.

\begin{figure}
\begin{center}
\begin{tabular}{c}
\includegraphics{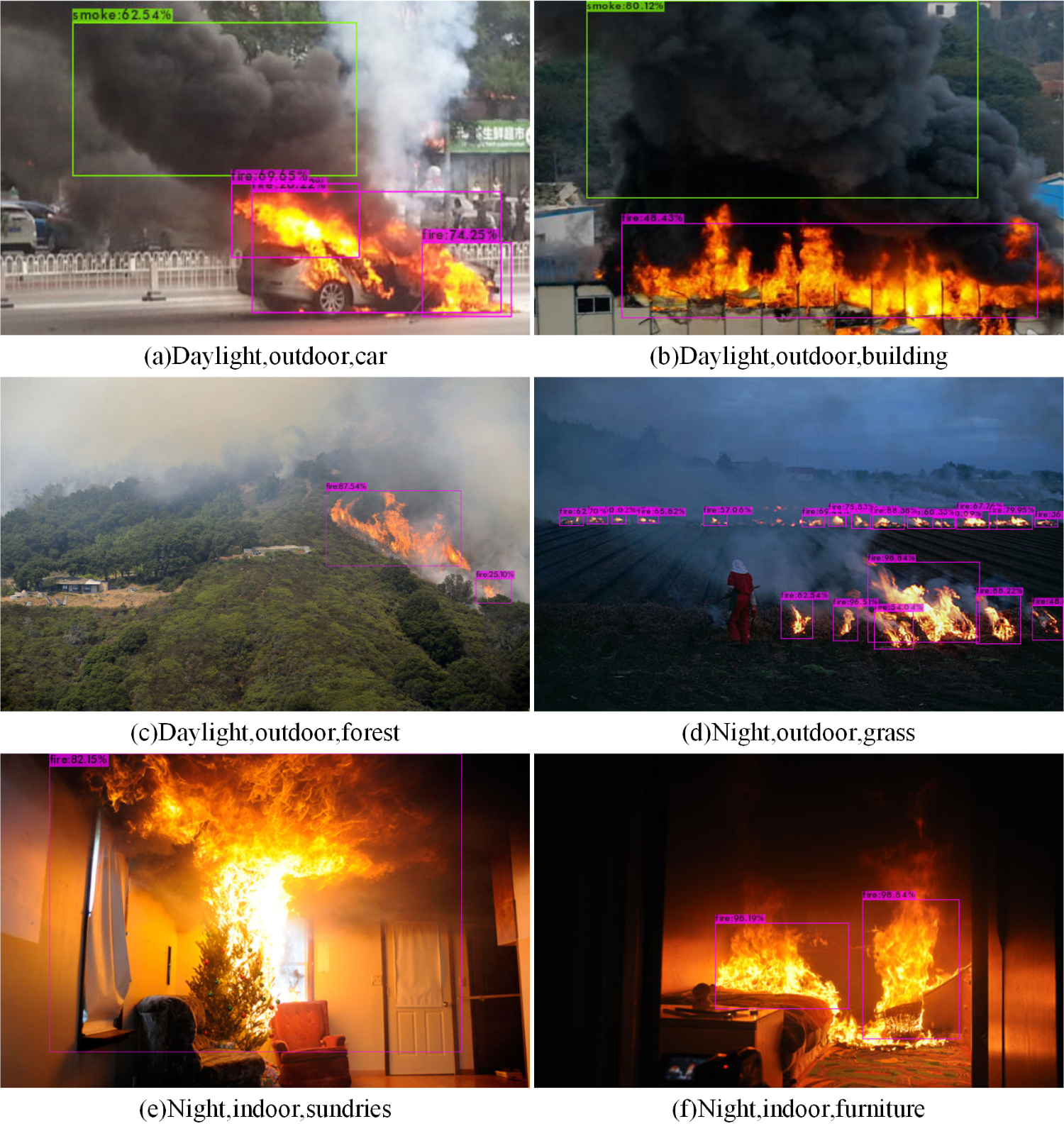}
\end{tabular}
\end{center}
\caption 
{ \label{fig:ex7}
Fire and smoke detection results On DS3 of FSDNet.(a)(b)Fire that often occur in cities. (c)(d)Fire that often occur in nature. (e)(f)Fire that often occur in indoor scenarios.} 
\end{figure} 

However, objects in fire detection tasks have uniqueness in that they undergo dynamic changes in morphology and color features over time, leading to corresponding changes in the overlap between the predicted object box and the true label box. For example, the true label box and the detected bounding box only partially overlap, but this does not have a significant impact on the detection performance; there may be multiple detected bounding boxes, but the true label box is only one.

The above two points can greatly affect the IOU between the predicted bounding box and the true label box, which in turn affects the algorithm's mAP. However, in the initial stage of fire detection, it is usually expected to determine whether there is a fire as quickly as possible, and in this stage, the relationship between the object box and the true box can be ignored, and the AP value is not calculated. Therefore, in this paper, the detection rate is used as an indicator to further verify the algorithm's performance by focusing on detecting the presence of fire. The specific detection results of FSDNet are presented to further demonstrate the reliability of the FSDNet detector.

The fire and smoke detection results of the FSDNet algorithm based on the MS-FS dataset are shown in Fig.~\ref{fig:ex7}, which further demonstrates the good performance of the FSDNet detector. (a)-(f) are some common fire scenarios in daily life, including both outdoor and indoor scenes, and all of them have certain detection difficulties. In scenes (b) and (f), the object shapes are varied and irregular; in scenes (c) and (e), there are significant differences in the proportion of the objects; in scene (d), there are many objects with significant differences in proportions. Nevertheless, FSDNet exhibits excellent performance in the specific categories, bounding box positions, and accuracy of detection.

\begin{figure}
\begin{center}
\begin{tabular}{c}
\includegraphics{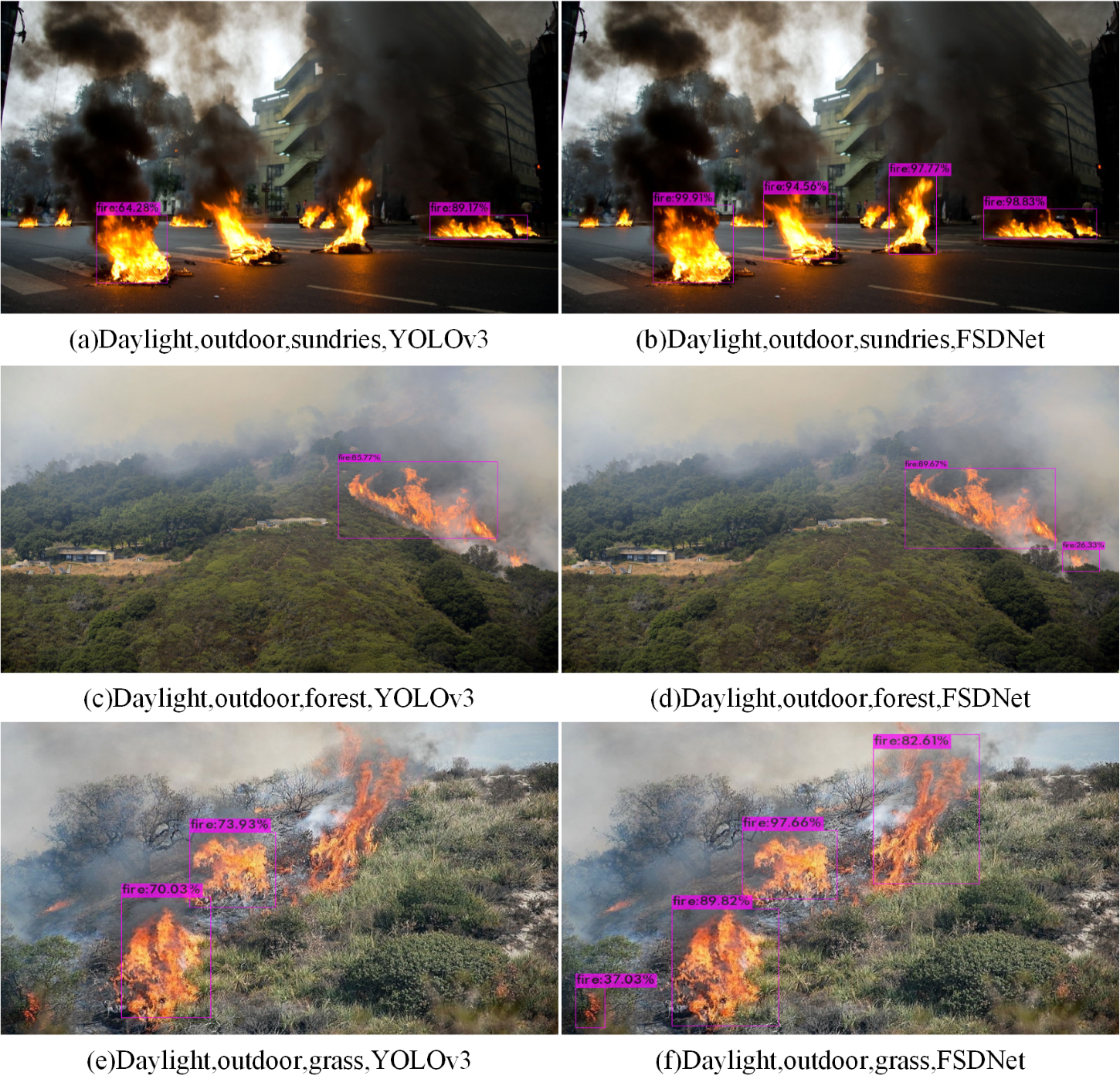}
\end{tabular}
\end{center}
\caption 
{ \label{fig:ex8}
Fire and smoke detection results comparison between YOLOv3 and FSDNet under same scenarios. The first and the second columns are the results of YOLOv3 and FSDNet, respectively.} 
\end{figure}

In order to more intuitively observe the fire detection performance of the FSDNet algorithm in complex scenes, this section compares the detection results of the FSDNet algorithm and the YOLOv3 algorithm on selected images. The specific results are shown in Fig.~\ref{fig:ex8}.

Fig.~\ref{fig:ex8} shows the comparison of the detection results of the FSDNet algorithm and the YOLOv3 algorithm on the same data. The left side is the detection result of the YOLOv3 algorithm, and the right side is the detection result of the FSDNet algorithm. There are 4 sets of scene comparisons.

For scene (a), the FSDNet algorithm detects more flame objects, and for the same flame objects, the predicted probability value is higher; for scene (b), when the same number of flame objects are detected, the FSDNet algorithm shows that More precise positioning accuracy and classification effect; for scenes (c) and (d), the FSDNet algorithm not only showed higher predicted probability values on the same object, but also detected smaller flame objects.

It can be clearly seen that compared with the YOLOv3 algorithm, the FSDNet 
algorithm can detect more and smaller flame objects in various scenarios, which is of great significance for fire detection and early fire monitoring in complex scenarios.

\section{Conclusions}
In this article, a fire detection network called FSDNet is proposed to improve detection accuracy while maintaining real-time performance. In the FSDNet network, a basic feature extraction module consisting solely of small convolutional kernels is further proposed, and a dense connection structure of convolutional layers is applied to strengthen feature propagation while reducing the number of parameters, aimed at enhancing the feature extraction ability of the backbone network and alleviating the problem of gradient disappearance. In addition, a spatial pyramid pooling structure and a Mosaic data augmentation method are introduced to improve the accuracy loss caused by the shape differences of flame objects and the imbalance of object proportions in the dataset, respectively. A new loss function is adopted to accelerate model training. Furthermore, a large-scale fire dataset covering diverse scenes, MS-FS, is independently constructed to provide strong data support.

Experimental results on the DS1, DS2, and MS-FS datasets demonstrate that FSDNet achieves higher classification accuracy than previous methods, and also outperforms YOLOv3 and high-performance YOLOv4 in detection accuracy, enabling precise and real-time fire detection. Overall, FSDNet proposed in this article is expected to have a promising application prospect in community fire monitoring and prevention.


\subsection* {Acknowledgments}
My mentor, Associate Professor Julie




\bibliography{report}   
\bibliographystyle{spiejour}   


\vspace{2ex}\noindent\textbf{First Author} is a postgraduate student majoring in communication engineering at Nanchang University. His research topics during his postgraduate period are deep learning and computer vision.

\listoffigures
\listoftables

\end{spacing}
\end{document}